\pdfoutput=1
\documentclass[10pt,twocolumn,letterpaper]{article}

\usepackage[pagenumbers]{cvpr} 

\usepackage{graphicx}
\usepackage{amsmath}
\usepackage{amssymb}
\usepackage{booktabs}
\usepackage{dblfloatfix}
\usepackage{csquotes}
\usepackage{multirow}
\usepackage[accsupp]{axessibility}
\usepackage[section]{placeins}

\usepackage[pagebackref=true,colorlinks=true,citecolor=blue]{hyperref}

\renewcommand*{\backref}[1]{}
\renewcommand*{\backrefalt}[4]{%
  \ifcase #1 %
No citations.
  \or
(p. #2).%
  \else
(p. #2).%
  \fi%
}
\newcommand\blfootnote[1]{%
  \begingroup
  \renewcommand\thefootnote{}\footnote{#1}%
  \addtocounter{footnote}{-1}%
  \endgroup
}

\usepackage[capitalize]{cleveref}
\crefname{section}{Sec.}{Secs.}
\Crefname{section}{Section}{Sections}
\Crefname{table}{Table}{Tables}
\crefname{table}{Tab.}{Tabs.}

\def\cvprPaperID{13} 
\def\confName{CVPR}
\def\confYear{2022}

\begin{document}

\title{CoRe: Color Regression for Multicolor Fashion Garments}

\author{Alexandre Ramé\\
\and
Arthur Douillard\\
\and
Charles Ollion\\
}

\author{Alexandre Ramé\textsuperscript{1,2}, Arthur Douillard\textsuperscript{1,2}, Charles Ollion\textsuperscript{2}\\
\textsuperscript{1}Sorbonne Université, \textsuperscript{2}Heuritech,
}

\maketitle

\begin{abstract}
Developing deep networks that analyze fashion garments has many real-world applications.
Among all fashion attributes, color is one of the most important yet challenging to detect.
Existing approaches are classification-based and thus cannot go beyond the list of discrete predefined color names.
In this paper, we handle color detection as a regression problem to predict the exact RGB values.
That's why in addition to a first color classifier, we include a second regression stage for refinement in our newly proposed architecture. 
This second step combines two attention models: the first depends on the type of clothing, the second depends on the color previously detected by the classifier. Our final prediction is the weighted spatial pooling over the image pixels RGB values, where the illumination has been corrected.
This architecture is modular and easily expanded to detect the RGBs of all colors in a multicolor garment.
In our experiments, we show the benefits of each component of our architecture.\blfootnote{Accepted at CVPR 2022 for the 5th Workshop on Computer Vision for Fashion, Art, and Design.}
\end{abstract}

\section{Introduction}

Convolutional Neural Networks (CNNs) \cite{krizhevsky2012imagenet, he2016deep} generated a lot of interest in the fashion industry. Recent datasets of fashion images \cite{liu2016deepfashion,zheng2018modanet,ge2019deepfashion2} encouraged various approaches for attributes classification \cite{bossard2012apparel,veit2015learning}, visual search \cite{yamaguchi2013paper,hadi2015buy,lynch2016images,corbiere2017leveraging} and object detection \cite{jia2018deep,rame2018omnia}.

\par One of the key attributes to describe a fashion item is its color. However, colors are subjective properties of garments as not all humans recognize colors the same way \cite{partos2016you}: their automatic estimation is therefore challenging. So far, most approaches consider the problem through the angle of \textbf{discrete color naming}: a classifier chooses amongst 11 colors of the English language \cite{berlin1991basic}, based on features from histograms \cite{baek2007vehicle,chen2014vehicle,van2009learning} or from CNNs \cite{wang2015color, rachmadi2015vehicle, networkpedestrian,yu2018weakly,yazici2018color,hickey2020hierarchical,8971326}. Recent approaches increased the number of color names up to 28 \cite{yu2018beyond} or 313 \cite{zhang2016colorful}. We take a step forward and tackle the problem through the angle of continuous color regression. Rather than only predicting an approximate discrete color name, we aim at predicting the exact \textbf{continuous color RGB}. This refined information is necessary for many industrial applications such as precise visual search and fine-grained trend detection to handle color inherent ambiguity. Moreover, the regression paradigm is more versatile and adaptable; this is especially important for real-world applications, where product requirements can evolve. In contrast, adding a new color name in classification approaches would require a dataset relabelling and the training of a new network.

\begin{figure}
\centering
\includegraphics[width=\linewidth]{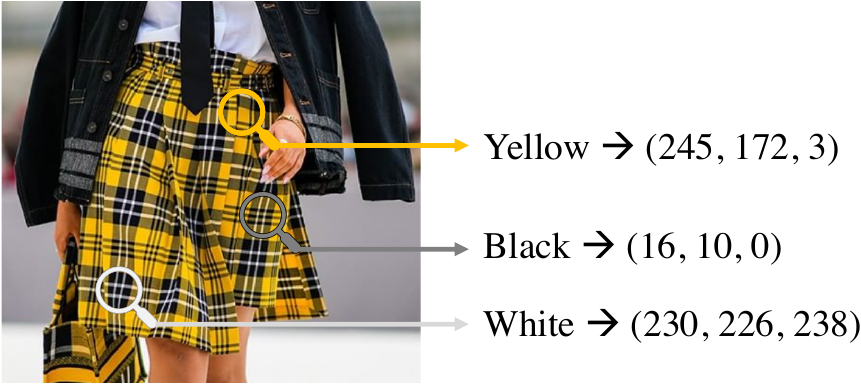}
\caption{\textbf{Two-stage color regression}. After detecting the main color names of the clothing (the skirt here), our architecture regresses the exact and refined RGB values.}
\label{fig:bigpicture}
\end{figure}%

To understand the challenge of this regression task, let's consider an unsupervised approach, that would naively average the different pixel RGBs of the image. That would fail for three reasons. \textit{First}, it would suffer from varying illuminations, sparking discrepancies between the raw pixels and what a human would perceive in ideal white light. \textit{Second}, it would consider `parasite' pixels from either the background or other garments. \textit{Third}, in case of multicolor garments, it would average pixels belonging to different colors and would predict neutral RGBs.

\paragraph{}
In this paper, we propose a new supervised two-stage architecture for color regression, inspired by the object detection literature \cite{ren2015faster}.
The first stage is a standard color classifier --- similar to previous works \cite{wang2015color,rachmadi2015vehicle, networkpedestrian,yu2018weakly,yazici2018color} --- which predicts the main colors of the considered garment. Our contributions lie in the second regression-based stage, refining the previous discrete prediction by weighting the different pixels of the illumination-corrected image.

\textit{First}, we developed an illumination module that automatically corrects the contrast of the image --- and this without supervision. This is inspired by the color constancy literature \cite{bianco2015color,barron2015convolutional,lou2015color,bianco2017single,shi2016deep,fourure2016mixed,hu2017fc4,sidorov2018artificial}.
\textit{Second}, in order to reduce the impact of complex backgrounds or complex garments structure, we leverage a semantic segmenter \cite{wieclawek2014car,networkpedestrian} pretrained to detect fashion garments. \textit{Third}, we detect all pixels in this garment that are close to the colors predicted by the first stage. This enables the handling of multicolor garments by removing pixels from conflicting colors. These last two components are combined to find the appropriate weighting of each pixel in the image --- enabling to focus on the appropriate regions of the image.

Empirically, we validate the effectiveness of our approach in our new fashion dataset collected from real-world images: we show that all our components contribute.
\section{Model}


\begin{figure*}
\centering
\includegraphics[width=0.8\textwidth]{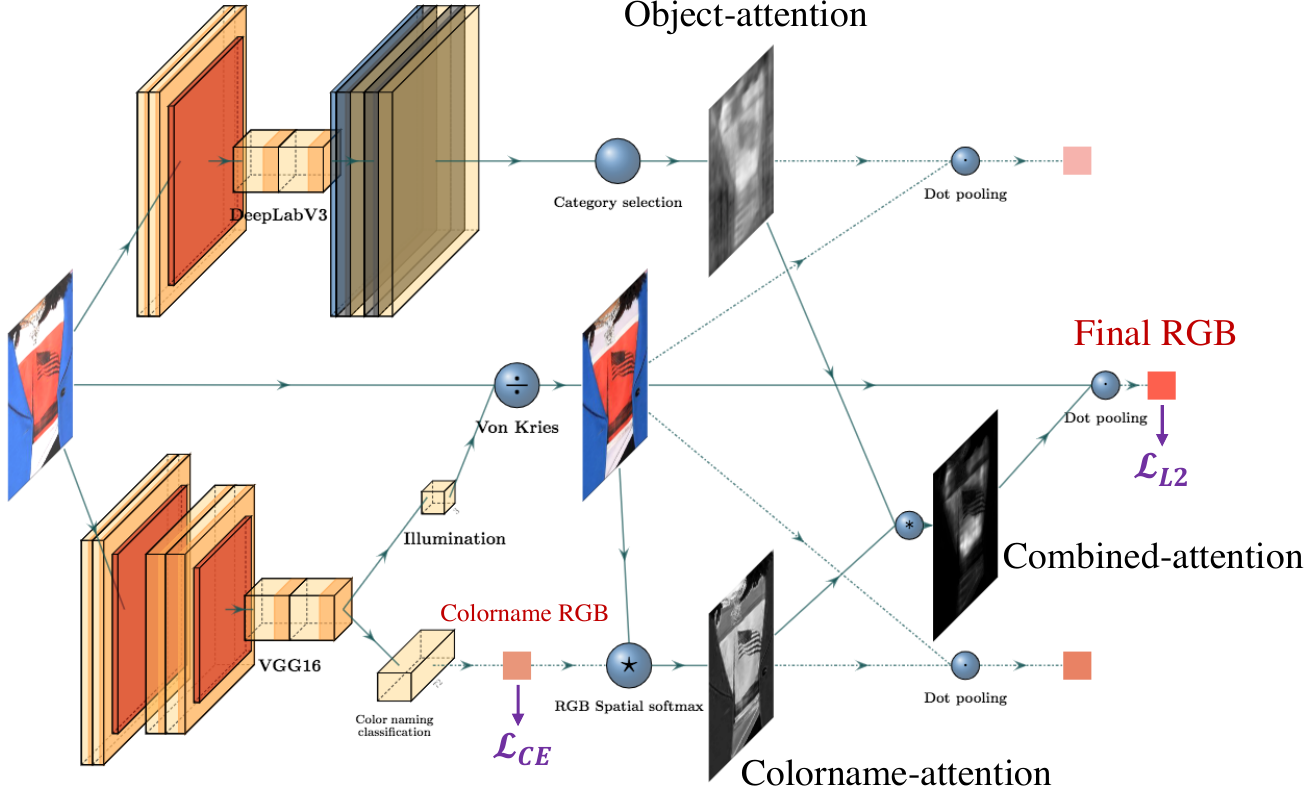}%
\caption{\textbf{Architecture components and training procedure}. \textit{First}, the \textit{DeepLabv3} \textbf{object-attention} network selects pixels inside the garment. \textit{Second}, the \textit{VGG16} CNN extract features before feeding two heads: the first predicts the image initial \textbf{illumination} (to be corrected with the \textit{Von Kries} method), and the second head detects the \textbf{main discrete color name} amongst 72 available colors (trained with a categorical cross entropy $\mathcal{L}_{CE}$). This main color is used to create a \textbf{colorname-attention}. These attentions are used for weighted spatial pooling over the pixels of the image for \textbf{color regression}.
Best results are obtained when colorname-attention and object-attention are combined, as this would select only pixels that are simultaneously inside the clothing and that have a RGB value close to the RGB of the predicted discrete color name. All components are trained so that the final RGB matches the RGB annotation with an Euclidean distance in the LAB space ($\mathcal{L}_{L2}$). This architecture is easily extended to detect the RGBs of multiple colors (red and white here).}
\label{fig:attentioncolorrgression}
\end{figure*}%

The network in Figure \ref{fig:attentioncolorrgression} is trained end-to-end on the task of continuous color regression, with annotated RGB values, but with neither illuminations nor clothing segmentation ground-truth annotations.

\subsection{Pixels correction}

Our model needs to correct the bias in the pixels of the image, for example the quality or the lighting of the photo. To do so, first, all images are \textbf{contrast normalized} in a preprocessing step by global histogram stretching in order to be robust across various lighting conditions. Note that this does not change the relative contributions of the three RGB channels. Second, to remove \textbf{illumination} color casts, we need to detect the initial illumination of the image (a scalar value per channel); this is a regression task, achieved easily by a deep neural network.
Note that following \cite{lou2015color}, we choose explicitly the \textit{VGG16} \cite{vgg16} neural architecture because of its capacity to extract low-level features \cite{Gatys2015ANA}.
Finally, each pixel in the image is corrected using the \textit{Von Kries} method \cite{von1902chromatic}, \textit{i.e.}, scaling each channel by the detected illumination value.

\subsection{Spatial pooling over pixels for color regression}

To sample only the relevant regions of the image, we apply a spatial pooling over the pixels of the image weighted by two complementary attention module. They modulate the importance of each pixel in the spatial pooling. The first \textbf{object-attention} focuses on the clothing category, the second \textbf{colorname-attention} focuses on the color name predicted. These attention masks are then multiplied pixelwise to create the \textbf{combined-attention}.

\subsubsection{Through object-attention}%

Following the work from \cite{yu2018weakly}, we use the popular fully convolutional \textit{Deeplabv3} \cite{chen2017rethinking}. This attention network is pretrained on the task of semantic segmentation over the clothing crops from \textit{Modanet} \cite{zheng2018modanet} and therefore has an emphasis on the garment surface: it produces an \textbf{object-attention} different for each clothing category (top, coat, \dots). This spatial prior is then fine-tuned: ideally, it would learn to identify the regions which contain the relevant color information given a garment type.

\subsubsection{Through colorname-attention}

The previous object-attention will fail for multicolor garments. Indeed, if the garment contains a set of distinct and distant RGB values, the mean of this set would be predicted: it leads to unsaturated color predictions. For example, in Figure \ref{fig:attentioncolorrgression}, including the white pixels from the shirt in our spatial pooling would predict a RGB closer to light pink.

\paragraph{Predicting the RGB of the main color}
For simplicity, we first describe how to detect only the RGB of the main color of a (potentially multicolor) garment.
\textit{First}, we detect the main color name; specifically, following previous works, the \textit{VGG16} features are followed by a fully connected layer with a softmax activation function that predicts a distribution over 72 color names. This is trained by minimizing a categorical cross-entropy loss $\mathcal{L}_{CE}$.

\textit{Second}, our \textbf{colorname-attention} only selects pixels in the image sufficiently close to the color detected by this first color classifier. We map this discrete color to its $\mathcal{RGB}_{c}$ continuous value: \textit{e.g.}, (134, 71, 71) for \textit{velvet red}. Now, given a pixel $p$ of RGB value $\mathcal{RGB}_{p}$, the colorname-attention $CA_{p}$ for pixel $p$ is:
\begin{equation}
CA_{p} \propto exp^{-\frac{1}{{127.5}^2 * \mathcal{T}}{(\mathcal{RGB}_{p} - \mathcal{RGB}_{c})}^2}.%
\label{eq:rgbspatialpooling}
\end{equation}%

It sums to $1$ over all pixels and its peakedness depends on $\mathcal{T}$, the temperature of the \textbf{RGB spatial softmax}. $\mathcal{T}$ is the only parameter of this colorname-attention module, which can be \textit{either} a predefined hyperparameter, \textit{either} learned, \textit{either} input-dependant and predicted with a fully connected layer from VGG features. This last option works best as shown in our Experiments from Section \ref{expe}.

\paragraph{Predicting the RGBs of all colors}

We can easily generalize our approach for detecting multiple RGBs. To do so, we predict multiple color names with a sigmoid \cite{yazici2018color} followed by a combination of categorical cross entropy and binary cross entropy as in \cite{rame2018omnia}. Multiple colors share the same object-attention but have different colorname-attentions: therefore we predict different RGBs with our combined-attention.%

\par
The challenge is then to know how many colors should be predicted. Following the work from \cite{yazici2018color}, we explicitly \textbf{learn the number of colors} in each garment (up to 3 different colors). Note that we apply class weights for handling unbalanced classes.

\par
Selecting the color names with maximum scores would often predict several times the same major color: for example a \textit{light red} and a \textit{dark red} would converge towards a \textit{medium red}. Thus, we apply a \textbf{non-maximum-suppression} algorithm to delete predicted RGBs too close to each other. In Figure \ref{fig:attentioncolorrgression}, the network's confusion between two different kinds of reds would have prevented from predicting white.

\subsection{Training overview}

The LAB colorspace best models perceptual distance in colorization \cite{zhang2016colorful,cheng2015deep}: thus, the chosen regression objective is $\mathcal{L}_{L2}$, the \textbf{Euclidean distance} between predicted and ground truth colors converted to LAB rather than RGB. Our two losses --- color naming $\mathcal{L}_{CE}$ and color regression $\mathcal{L}_{L2}$  --- are summed before backpropagation; this enables the learning of all architecture components end-to-end.

\subsection{Discussion: analogy with Faster R-CNN}
Our two-stage approach is highly inspired by the \textit{Faster R-CNN} \cite{ren2015faster} architecture. The anchors of the Region Proposal Network are replaced by color names. In both cases, the first classifier gives a rough estimate, refined by the second regression stage. Specifically, given a selected anchor (resp. color name), the final regression adds a small continuous offset leading to a more precise box (resp. RGB values). Non-maximum-suppression algorithms are also standard to handle overlapping crops in object detection.

\section{Experiments}%
\label{expe}

\subsection{Setup}

\begin{table*}[!t]%
    \caption{\textbf{Results for RGB regressions}: percentage of predictions closer to the ground truths RGBs than several thresholds ($10$, $20$, $30$, $40$), according to the \textit{deltaE ciede2000} distance \cite{fairchild2013color,mokrzycki2011colour}. \textbf{Bold} highlights best score.}%
    \centering
    \resizebox{\textwidth}{!}{%
        \begin{tabular}{c| c c c c|| c c c c| c c c c}%
            \hline
            \multicolumn{5}{c||}{Method}                  & \multicolumn{4}{c|}{Main color} & \multicolumn{4}{c}{All colors}                                                                                                                                                \\
            \hline
            Name                                          & Color-attention                 & Object-attention                    & Illumination & $\mathcal{T}$ & $\leq 10$   & $\leq 20$   & $\leq 30$   & $\leq 40$   & $\leq 10$   & $\leq 20$   & $\leq 30$   & $\leq 40$   \\
            \hline
            Unsupervised K-means Clustering \cite{kmeans} &                                 &                                     &              &               & 47          & 72          & 83          & 89          & 19          & 31          & 37          & 42          \\
            Colorname RGB \cite{yazici2018color} &                                 &                                     &              &               & 51          & 72          & 87          & 92          & 34          & 52          & 64          & 68          \\
            \hline
            Ours                                          &                                 & \checkmark                          &              & -             & 50          & 78          & 90          & 95          & -           & -           & -           & -           \\
            Ours                                          & \checkmark                      &                                     &              & Set to $1$    & 48          & 76          & 88          & 93          & 35          & 56          & 65          & 68          \\
            Ours                                          & \checkmark                      & \checkmark                          &              & Set to $1$    & 59          & 87          & 93          & 95          & 44          & 64          & 69          & 71          \\
            Ours                                          & \checkmark                      & \checkmark                          &              & Trainable     & 62          & 87          & 93          & 95          & 46          & 65          & 69          & 71          \\
            Ours                                          & \checkmark                      & \checkmark                          & \checkmark   & Trainable     & 67          & 89          & \textbf{94} & \textbf{96} & 48          & 66          & 70          & 72          \\
            Our best                                      & \checkmark                      & \checkmark                          & \checkmark   & Predicted     & \textbf{73} & \textbf{90} & 93          & 95          & \textbf{54} & \textbf{68} & \textbf{71} & \textbf{73} \\
            \hline
        \end{tabular}
    }%
    \label{table:rgbexperiments}
\end{table*}
\paragraph{Dataset}
As far as we know, there is no dataset for the task of continuous colors regression. Thus, we have collected a new dataset of 30,269 fashion garments: 5,363 coats, 8,166 dresses, 3,991 pants, 6,871 shoes and 5,878 tops. The validation dataset and the test dataset are composed of 2000 images: the other images are used for training. Each garment was labeled by a single operator with its exact color RGB: not like it appears in the image, but like the operator thought it would appear in ideal white light. The annotation process is therefore more complex and time-consuming than classical classification. The color names can automatically be derived by nearest neighbors in the \textit{LAB} space. In case of multicolor items, the operators were asked to tag them in decreasing order of importance.

\paragraph{Implementation}%
Our code is in Tensorflow \cite{tensorflow2015-whitepaper} and Keras \cite{chollet2015keras}. We chose an Adam \cite{kingma:adam} optimizer with a learning rate of 0.0001 and a batch of 16 images during 50 epochs. We applied standard data augmentation methods: random cropping and translation.

\subsection{Results}%
\begin{figure}[h]
\centering
\includegraphics[width=1.0\linewidth]{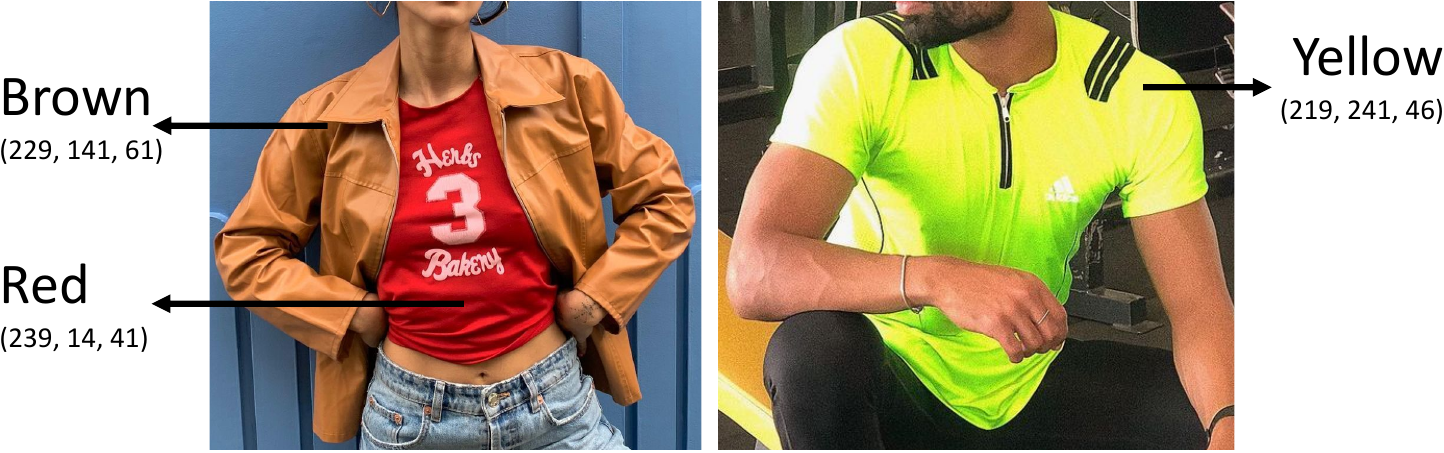}
\caption{\textbf{Visualization of the predictions}. 
This picture illustrates predictions on two images. On the left image, the two interleaved garments (jacket and t-shirt) are well predicted despite their closeness thanks to our attention mechanism. In the right image, our model distinguishes the good yellow RGB despite the illumination and shadows that harden the task.}
\label{fig:preds}
\end{figure}

Table \ref{table:rgbexperiments} summarizes our experiments. To compare approaches, we count the percentage of predictions that are closer  to the color annotation than a given threshold, leveraging the \textit{deltaE ciede2000} distance \cite{fairchild2013color,mokrzycki2011colour} that is arguably \enquote{the best metric for understanding how the human eye perceives color difference} \cite{deltae}.
With our best network that incorporates all our components, 73$\%$ of all predictions have a distance to the main color smaller than 10; in practice, these differences are small to the human eye. We also report similar metrics when the goal is to detect all colors in the garment. We showcase in Figure \ref{fig:preds} various predictions.

\paragraph{Baselines}
We compare our model against two existing baselines. First we train an unsupervised \textit{K-means} pixels clustering \cite{kmeans} directly on the pixel space, after a background removal done by an external semantic segmentation model trained on Modanet \cite{zheng2018modanet}. Colors are ranked according to the number of pixels in each cluster. The second baseline, \textit{Colorname RGB}, is the direct extension of the multitagger approach from \cite{yazici2018color}: it detects the color names (\textit{e.g.} \textit{velvet red}, \textit{dark purple}, etc.) and then produces the associated RGB values. This baseline is depicted in Figure \ref{fig:attentioncolorrgression}.

\paragraph{Attention}
First, fine-tuning the semantic segmentation model on the regression task (object-attention) already surpasses previous approaches when predicting only the main color. The colorname-attention improves results in the multi-color setup. These two attentions, when combined, are mutually reinforcing and complementary.

\paragraph{Temperature}
We show that the temperature $\mathcal{T}$ value is important: bigger $\mathcal{T}$ leads to sharp distribution and takes into account fewer pixels from the initial images that with a lower $\mathcal{T}$. Rather than grid-searching its optimal value, it can be learned for improved results. Moreover, the optimal $\mathcal{T}$ depends on the image: therefore, best results are obtained with $\mathcal{T}$ predicted from VGG features. This analysis is consistent with recent insights for calibration via input-dependant temperature scaling \cite{zhang2020mix}.

\paragraph{Illumination}
Finally, including the illumination module improves performances. Future work could further improve results by pretraining on the \textit{Color Checker Dataset} \cite{gehler2008bayesian}.

\section{Conclusion}

In this paper we addressed the color regression problem for fashion garments. By collecting a unique dataset of 30,269 images, we empirically show the benefits of our newly proposed two-stage architecture. These performance gains would have a potential impact on many real-world usages, notably to better detect fashion trends and for visual search. Finally, we hope to shed light on properties of fashion garments (perhaps surprisingly) complex to detect.

\newpage
{\small
\bibliographystyle{ieee_fullname}
\bibliography{egbib}
}

\end{document}